\documentclass[10pt,twocolumn,letterpaper]{article}

\usepackage{cvpr}
\usepackage{times}
\usepackage{epsfig}
\usepackage{graphicx}
\usepackage{amsmath}
\usepackage{amssymb}
\usepackage{algorithm}
\usepackage{algorithmicx}
\usepackage[noend]{algpseudocode}

\usepackage[pagebackref=true,breaklinks=true,letterpaper=true,colorlinks,bookmarks=false]{hyperref}

\cvprfinalcopy % *** Uncomment this line for the final submission

\begin{document}
	
	%%%%%%%%% TITLE
	\title{Video Object Segmentation with Re-identification}
	
	\author{Xiaoxiao Li$^{1}$ \quad Yuankai Qi$^{2}$ \quad Zhe Wang$^{3}$ \quad Kai Chen$^{1}$ \quad Ziwei Liu$^{1}$ \quad Jianping Shi$^{3}$ \quad \\
		\quad Ping Luo$^{1}$ \quad Xiaoou Tang$^{1}$ \quad Chen Change Loy$^{1}$ \\
		\and
		% $^1$\small{Department of Information Engineering, The Chinese University of Hong Kong}\\
		% $^2$\small{Department of Computer Science and Technology, Harbin Institute of Technology}\\
		% $^3$\small{Sensetime Group Limited}\\
		$^1$The Chinese University of Hong Kong \\
		\and
		$^2$Harbin Institute of Technology \\
		\and
		$^3$SenseTime Group Limited \\
		%{\tt\small \{lx015,lz013,pluo,ccloy,xtang\}@ie.cuhk.edu.hk}
	}
	
	\maketitle
	%\thispagestyle{empty}
	
	%%%%%%%%% ABSTRACT
	\begin{abstract}

Conventional video segmentation methods often rely on temporal continuity to propagate masks.
Such an assumption suffers from issues like drifting and inability to handle large displacement.
To overcome these issues, we formulate an effective mechanism to prevent the target from being lost via adaptive object re-identification.
Specifically, our Video Object Segmentation with Re-identification (VS-ReID) model includes a mask propagation module and a ReID module.
The former module produces an initial probability map by flow warping while the latter module retrieves missing instances by adaptive matching.
With these two modules iteratively applied, our VS-ReID records a global mean (Region Jaccard and Boundary F measure) of 0.699, the best performance in 2017 DAVIS Challenge.

\end{abstract}

	%%%%%%%%% INTRODUCTION
	
\section{Introduction}

Video object segmentation in 2017 DAVIS Challenge \cite{Pont-Tuset_arXiv_2017} is non-trivial -- a video typically consists of more than one annotated object, with many distractors, small objects and fine structures. The complexity of the problem increases with severe inter-object occlusions and fast motion.
	
Conventional approaches that rely on temporal continuity suffer from issues like drifting and inability to handle large displacement. To overcome these issues, we formulate an effective mechanism to prevent the target from being lost via adaptive object re-identification. Specifically, our Video Object Segmentation with Re-identification (VS-ReID) model includes a mask propagation module and a ReID module. The mask propagation module is a two-stream convolutional neural network, inspired by \cite{Perazzi2017}. The RGB branch of the mask propagation module accepts a bounding box and a guided probability map as input, and produces a segmentation mask for the main instance as an output. The guided probability map is obtained from adjacent frames' predictions by flow warping. In addition to the RGB branch, we also train an optical flow branch to incorporate the temporal information. The final segmentation mask of the image patch is obtained by averaging the predictions of these two branches. 

To cope with frequent occlusions and large pose variations in dynamic scenes, we leverage object re-identification module to retrieve missing instances. Specifically, when missing instances are re-identified with a high confidence, they are assigned with a higher priority to be recovered during the mask propagation process. For each retrieved instance, we take its frame as the starting point and use the mask propagation module to bi-directionally generate the probability maps in its adjacent frames. 

With the updated probability maps, the mask propagation module and ReID module of VS-ReID are iteratively applied to the whole video sequence until no more high confidence instances can be found.  Finally, for each frame, the instance segmentation results  are obtained by merging the probability maps of all the instances. With both flow warping to ensure temporal continuity and object re-identification to recover missing objects, VS-ReID records a global mean (Region Jaccard and Boundary F measure) of 0.699, the best performance in 2017 DAVIS Challenge.

	%-------------------------------------------------------------------------

	%%%%%%%%% RELATED WORK
	\section{Related Work}

The realm of object segmentation witnesses drastic progress these days, including the marriage of deep learning and graphical models~\cite{zheng2015conditional, liu2015semantic} and the efforts to enable real-time inference on high-res images~\cite{li2017not, zhao2017icnet}.
Since most visual sensory data are videos, it is crucial to extend object segmentation from image to video.
Existing video segmentation methods~\cite{liu2016deep, Perazzi2017} rely on temporal continuity to establish spatio-temporal correlation.
However, real-life videos exhibit severe deformation and occlusion, rendering such assumption to suffer from issues like drifting and inability to handle large object  displacement.
In this work, we propose a novel method known as Video Object Segmentation with Re-identification (VS-ReID) to overcome these issues.

% The realm of object segmentation witnesses drastic progress these days, including the marriage of deep learning and graphical models~\cite{}, the integration of multi-level information~\cite{} and the efforts to enable real-time inference~\cite{}.
% As the results of semantic image segmentation gradually become accurate, there are two notable research trends emerging, both of which constitute the building elements for DAVIS video segmentation challenge:

% \noindent
% \textbf{Video Segmentation.}
% %
% Most of visual sensory data are videos, thus it is crucial to extend object segmentation from image to video.
% However, since real-life videos observe lots of deformation and occlusion, which are not well handled by the existing methods.

% \noindent
% \textbf{Instance Segmentation.}
% %
% Instance segmentation aims at segmenting out each instance in the image, either class-aware or class-agnostic.
% It can be viewed as the combination of object detection and semantic segmentation.

	%-------------------------------------------------------------------------

	%%%%%%%%% APPROACH
	\section{Approach}
Our VS-ReID model includes a mask propagation module and a re-identification module.
The mask propagation module propagates the probability map from the predicted frame to the adjacent frames.
Meanwhile, we employ the re-identification module to retrieve instances that are missing during the mask propagation process.
Two modules are iteratively applied to the whole video sequence.
Next, we will first present these two modules respectively in Sec.~\ref{sec:mask_propagation_module} and Sec.~\ref{sec:re_identification_module}, then introduce the algorithm of VS-ReID in Sec.~\ref{sec:VS-ReID}.

\begin{algorithm}
	\caption{Mask propagation for single object}   
	\label{alg:mask_propagation}
	\begin{algorithmic}[1]
		\small
		\Procedure{$\mathcal{M}_{mp}$}{$I_i, I_j, P_{i, k}$}
%		\Require two adjacent frames, $\{I_i, I_j\}$; the probability map for instance $k$ in the frame $i$, $P_{i, k}$;
%		\Ensure the probability map for instance $k$ in the frame $j$, $P_{j,k}$;
		\State $P_{j,k} \gets 0$ \Comment{initialize}
		\State $f_{i \to j} \gets \mathcal{F}(I_i, I_j)$ \Comment{extract the optical flow}
%		\label{ code:fram:extract }
		\State $P_{i \to j, k} \gets \mathcal{W}(P_{i, k}, f_{i \to j})$ \Comment{flow guided warp}
		\State $b \gets \mathrm{Box}(P_{i \to j, k} > 0.5)$ \Comment{obtain the bounding box}
		\State $P_{j, k}^b \gets \mathcal{N}_{mp}(I_{j}^b, f_{i \to j}^b, P_{i \to j, k}^b)$
		\State \textbf{return} $P_{j, k}$
		\EndProcedure
	\end{algorithmic}  
\end{algorithm}

\subsection{Mask Propagation Module}
\label{sec:mask_propagation_module}
The inference algorithm of mask propagation is summarized in Algorithm~\ref{alg:mask_propagation}. 
Given two adjacent frames $I_i, I_j$, and the pixel-level probability map for instance $k$ in the frame $i$, $P_{i,k}$, we aim to predict the probability map for instance $k$ in the frame $j$, $P_{j,k}$.

Following~\cite{Perazzi2017, khoreva_lucid_dreams17}, we first obtain the coarse estimation of $P_{j,k}$, $P_{i \to j, k}$, from $P_{i,k}$ by flow guided warping.
We use FlowNet2.0~\cite{IMKDB17} to extract the optical flow $f_{i \to j}$ between frame $i$ and $j$.
The probability map $P_{i,k}$ is warped to $P_{i \to j, k}$ according to $f_{i \to j}$ by a bilinear warping function $\mathcal{W}$.
After that, we employ a convolutional neural network, mask propagation network $\mathcal{N}_{mp}$, to further refine the coarse estimation.
 Rather than full-resolution images as in~\cite{Perazzi2017, khoreva_lucid_dreams17}, our mask propagation network accepts size-normalized patches that enclose objects of interest as input and produces the refined probability patch. Using object patches as input allows our model to better cope with objects of different scales.
More specifically, we crop the patches $ I_j^b$, $f_{i \to j}^b$ and $P_{i \to j, k}^b$ from full-image by instance bounding box $b$.
Then we resize those patches into a fixed size and feed them into the mask propagation network to get the probability patch $P_{j,k}^b$.
Finally, $P_{j,k}^b$ is resized back to the original size, and fill into a full size zero map to generate the prediction of $P_{j,k}$.
Unlike full-image based network~\cite{Perazzi2017, khoreva_lucid_dreams17}, our method can easily capture the small objects and fine structures.

\begin{figure}
	\centering
	\includegraphics[width=0.48\textwidth]{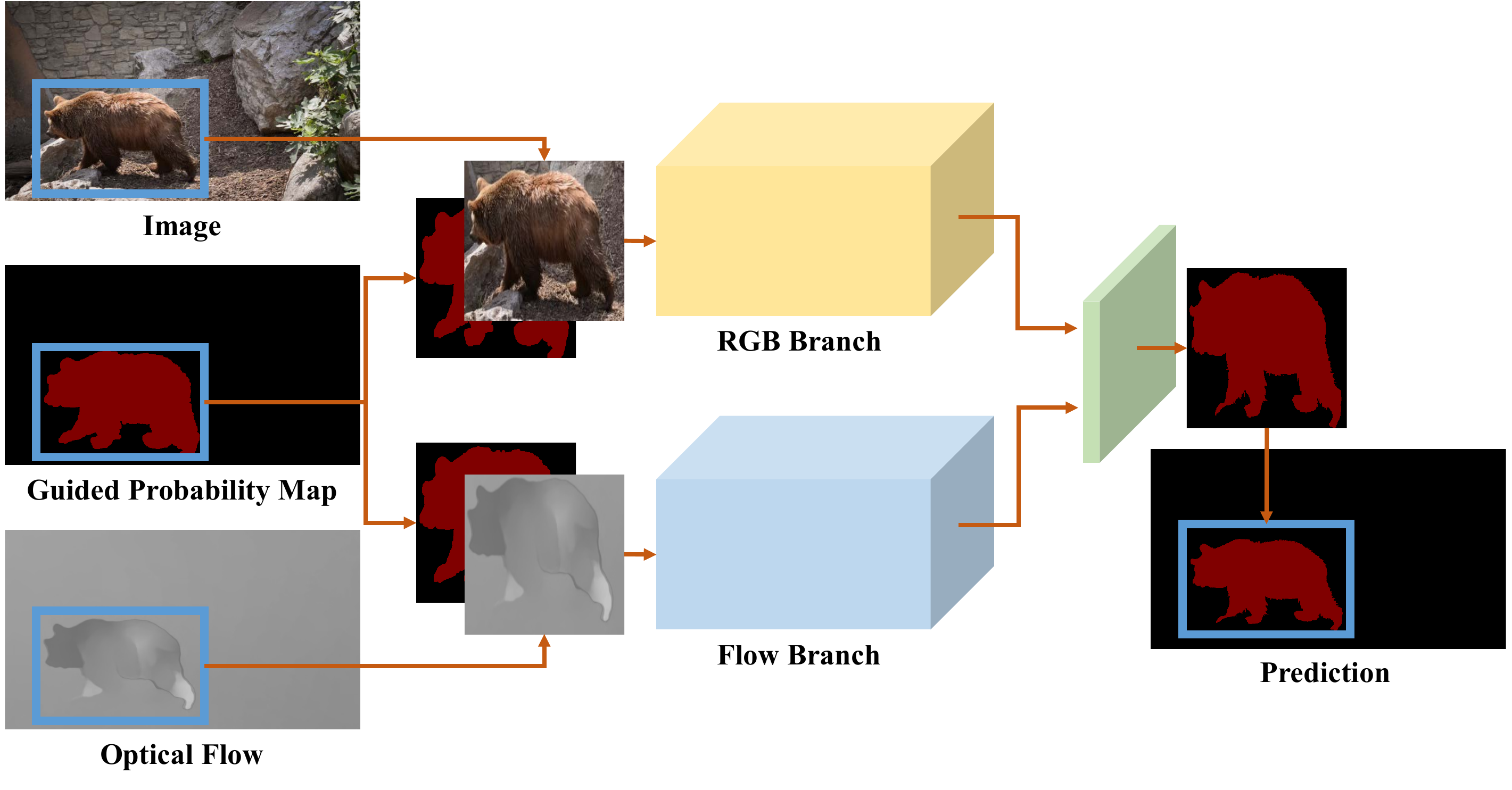}
	\caption{\small{Network architecture of mask propagation network. \textbf{Best viewed in color.}}}
	\label{fig:mp_network}
	\vspace{-12pt}
\end{figure}

\begin{figure*}[t]
	\centering
	\includegraphics[width=0.95\textwidth]{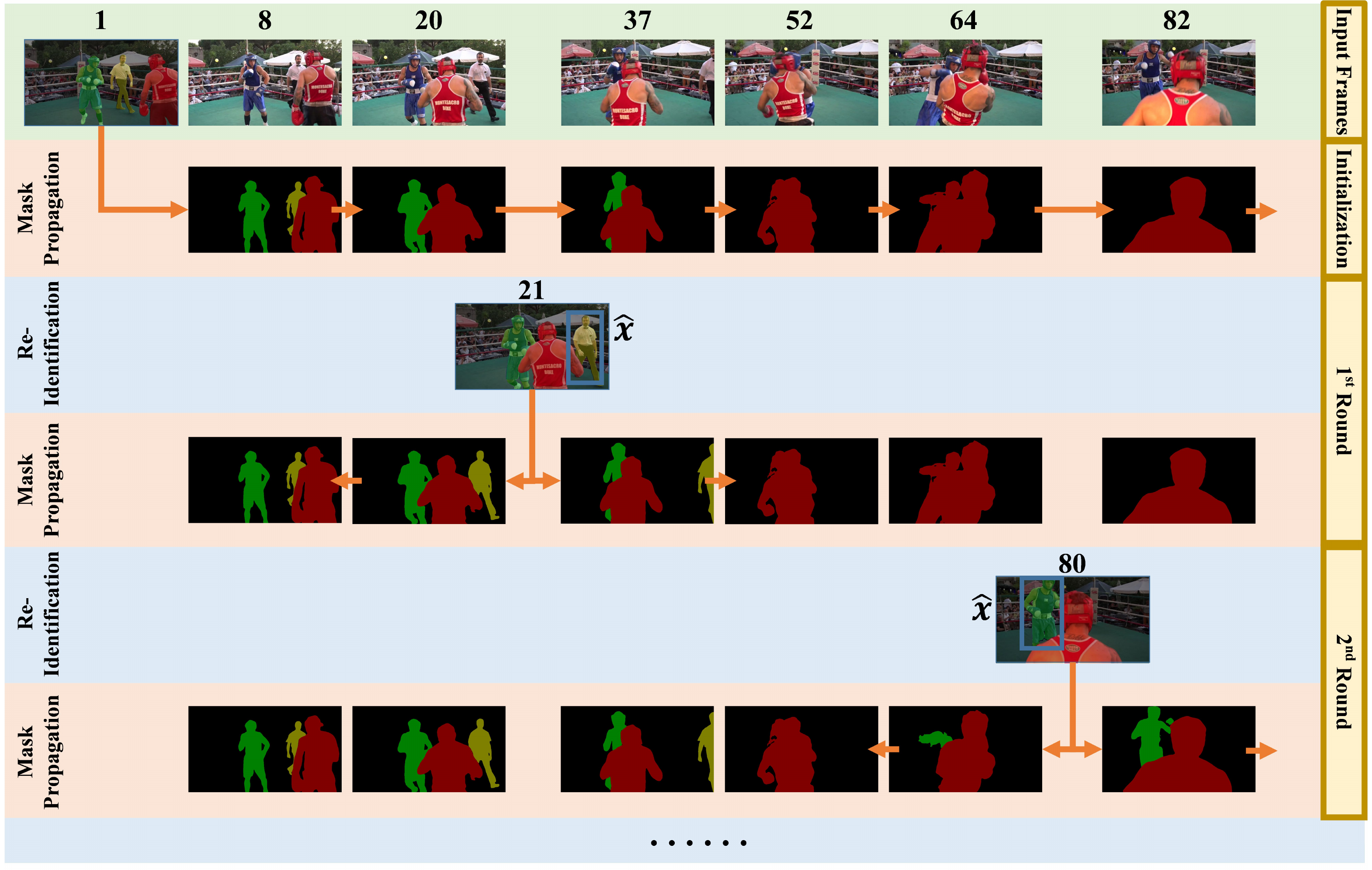}
	\caption{\small{Pipeline of our Video Object Segmentation with Re-identification (VS-ReID) model. \textbf{Best viewed in color.}}}
	\label{fig:pipeline}
	\vspace{-10pt}
\end{figure*}

\noindent
\textbf{Mask Propagation Network.} 
As shown in Fig.~\ref{fig:mp_network}, our mask propagation network is a two-stream convolutional neural network, inspired by~\cite{khoreva_lucid_dreams17}. 
However, several important modifications are necessary to further improve the network performance.
First, we adopt the much deeper ResNet-101~\cite{He2015} network to increase the model capacity.
Second, as we mentioned before, since our network takes patches as input, it is capable of capturing more details compared with full-image based network.
We also slightly enlarge the bounding box to keep more contextual information.
Third, to increase the resolution of prediction, we enlarge the size of feature maps by decreasing the convolutional stride and replace convolutions by dilated convolutions.
Similar to \cite{chen2014semantic}, atrous spatial pyramid pooling and multi-scale testing are also employed.
Last but not least, after independent branch training, two streams are jointly fine-tuned to further improves the performance.

\begin{algorithm} 
	\caption{Re-identification module}   
	\label{alg:reid}
	\begin{algorithmic}[1]
		\small
		\Procedure{$\mathcal{M}_{reid}$}{$I_i, P_{i, k}, t_k$}
%		\REQUIRE a video frame, $I_i$; the pixel-level probability map for instance $k$ in the frame $i$, $P_{i,k}$; and the template of instance $k$, $t_k$;
%		\ENSURE retrieve boundary box of the instance $k$ in the frame $i$, $x$ and corresponding re-identification score $s$;
			\State $X \gets \mathcal{N}_{det}(I_i)$ \Comment{obtain the candidate boxes}
	
			\For {$x_j \in X$}
				\State $s_j \gets S_C(\mathcal{N}_{reid}(I_i^{x_j}), \mathcal{N}_{reid}(t_k))$ \Comment{$S_C$ denotes the cosine similarity}
			\EndFor
			
			\State $\hat{j} \gets \arg\underset{j \le \left\vert{X}\right\vert }{\max}\, s_j$
			\State $b \gets \mathrm{Box}(P_{i,k} > 0.5)$
			\If{$s_{\hat{j}} > \rho_{reid}$ \textbf{and} $\mathrm{IoU}(x_{\hat{j}}, b)< \rho_{occ}$}
				\State \textbf{return} $x_{\hat{j}}, s_{\hat{j}}$
			\Else
				\State \textbf{return} $x_{\hat{j}}, -1$ \Comment{fail or unnecessary}
			\EndIf
%		$s_{\hat{j}} - \underset{l \ne k}{\max}\, S_C( \mathcal{N}_{reid}(I_i^{x_{\hat{j}}}), \mathcal{N}_{reid}(t_l)) > \rho_{diff}$ \textbf{and}
%		\IF {$s_{\hat{j}} > \rho_{reid}$ and $IoU(x_{\hat{j}}, b)< \rho_{occ}$}
%		\RETURN $x_{\hat{j}}, s_{\hat{j}}$
%		\ELSE
%		\RETURN $x_{\hat{j}}, -1$ \Comment{fail or unnecessary}
%		\ENDIF
		\EndProcedure
	\end{algorithmic}  
\end{algorithm} 

\subsection{Re-identification Module}
\label{sec:re_identification_module}
Our mask propagation module is based on the short-term memory and it highly relies on temporal continuity.
However, frequent occlusions and large pose variations are very common in dynamic scenes and likely to cause failures in mask propagation.
To overcome these issues, we leverage object re-identification module to retrieve missing instances.
Re-identification module incorporates long-term memory, which complements mask propagation module and makes our system more robust.

As summarized in Algorithm~\ref{alg:reid}, during the iterative refinement in VS-ReID, our re-identification module takes a single frame $I_i$, the current pixel-level probability map $P_{i, k}$ which is predicted in the previous round of inference for instance $k$ in the frame $i$, and the template of instance $k$, $t_k$ as input, produces the retrieved boundary box $x$, and corresponding re-identification score $s$.
In this module, we obtain the candidate bounding boxes $X$ in frame $I_i$ through a detection network $\mathcal{N}_{det}$.
For each candidate bounding box $x_j$, the re-identification score between $x_j$ and $t_k$ is conducted through measuring the cosine similarity between their features that are extracted from a re-identification network $\mathcal{N}_{reid}$.
Suppose $x_{\hat{j}}$ is the most similar candidate bounding box, it is only accepted as the final result if two conditions are satisfied:
First, $x_{\hat{j}}$ is sufficiently similar with the template $t_k$, that is, the re-identification score between $x_{\hat{j}}$ and $t_k$ is larger than a threshold $\rho_{reid}$;
Second, current $P_{i, k}$ is not consistent with $x_{\hat{j}}$, otherwise we do not need to retrieve the instance k in frame i.
To evaluate this condition, we compute the IoU score between $x_{\hat{j}}$ and current bounding box from $P_{i, k}$.
If it is less than another threshold $\rho_{occ}$, which means they are inconsistent, we believe that we have made a wrong prediction of $P_{i, k}$ in the previous rounds and accept $x_{\hat{j}}$ as the retrieve bounding box.
Those two thresholds are selected on the validation set.

\noindent
\textbf{Detection \& Re-identification Network.}
We directly adopt the Faster R-CNN~\cite{renNIPS15fasterrcnn} as our detection network $\mathcal{N}_{det}$.
For the re-identification network $\mathcal{N}_{reid}$, we employ the architecture of `Identification Net' in~\cite{xiaoli2017joint} and retrain this network for the general object re-identification task.

\begin{algorithm}
	\caption{VS-ReID algorithm}   
	\label{alg:VS-ReID}
	\algnewcommand\algorithmicto{\textbf{to}}
	\algnewcommand\algorithmicdownto{\textbf{downto}}
	\algrenewtext{For}[3]%
	{\algorithmicfor\ #1 = #2 \algorithmicto\ #3 \algorithmicdo}
	\algrenewtext{While}[3]%
	{\algorithmicfor\ #1 = #2 \algorithmicdownto\ #3 \algorithmicdo}
	\begin{algorithmic}[1]
		\small
		\Procedure {VS-ReID}{$\{I\}, \{P_{1}\}$}
%		\REQUIRE video frames, $\{I\}$, number of instances $K$, ground-truth probability map in the first frame for all instances $\{P_{0}\}$;
%		\ENSURE all frames' probability maps $\{P\}$;
			\State $N \gets \left\vert\{I\}\right\vert$	\Comment{number of frames}
			\State $K \gets \left\vert\{P_{1}\}\right\vert$ \Comment{number of instances}
			\For{$k$}{1}{$K$}
				\State obtain the template $t_k$ from $P_{1,k}$
			\EndFor

			\For{$i$}{2}{$N$}						\Comment{initialize probability maps}
				\For{$k$}{1}{$K$}
					\State $P_{i,k} \gets \mathcal{M}_{mp}(I_{i-1}, I_{i}, P_{i-1,k})$
					\State $c_{i,k} \gets 1$
				\EndFor
			\EndFor
			\Loop
				\State $\hat{s} \gets -1$
				\For{$i$}{2}{$N$}	\Comment{retrieve instances}
					\For{$k$}{1}{$K$}
						\State $x, s \gets \mathcal{M}_{reid}(I_{i}, P_{i,k}, t_k)$
						\If {$s > \hat{s}$ \textbf{and} $c_{i,k} \ne i$}
							\State $\hat{s} \gets s, \hat{x} \gets x, \hat{i} \gets i, \hat{k} \gets k$
						\EndIf
					\EndFor
				\EndFor
				\If {$\hat{s} < 0$}
					\State \textbf{break} \Comment{no instance retrieved}
				\Else 
					\State $P_{\hat{i},\hat{k}} \gets 0, f_{\hat{i}} \gets \mathcal{F}(I_{\hat{i}}, I_{\hat{i}+1})$
%					\State $f_{\hat{i}} \gets \mathcal{F}(I_{\hat{i}}, I_{\hat{i}+1})$
					\State $b \gets \mathrm{Box}(P_{1, \hat{k}} > 0.5)$
					\State $P_{\hat{i},\hat{k}}^{\hat{x}} \gets \mathcal{N}_{mp}(I_{\hat{i}}^{\hat{x}}, f_{\hat{i}}^{\hat{x}}, P_{1, \hat{k}}^{b})$ \Comment{recover}
					\For{$i$}{$\hat{i}+1$}{$N$} \Comment{forward propagate}
						\If {$|c_{i,\hat{k}} - i| > |\hat{i} - i|$}
							\State $P_{i,\hat{k}} \gets \mathcal{M}_{mp}(I_{i-1}, I_{i}, P_{i-1,\hat{k}})$
							\State $c_{i,\hat{k}} \gets \hat{i}$
						\EndIf
					\EndFor
					\While{$i$}{$\hat{i}-1$}{$2$} \Comment{backward propagate}
						\If {$|c_{i,\hat{k}} - i| > |\hat{i} - i|$}
							\State $P_{i,\hat{k}} \gets \mathcal{M}_{mp}(I_{i+1}, I_{i}, P_{i+1,\hat{k}})$
							\State $c_{i,\hat{k}} \gets \hat{i}$
						\EndIf
					\EndWhile
				\EndIf
			\EndLoop
			\State \textbf{return}  $\{P\}$
		\EndProcedure
	\end{algorithmic}  
\end{algorithm} 

\begin{figure}
	\centering
	\includegraphics[width=0.48\textwidth]{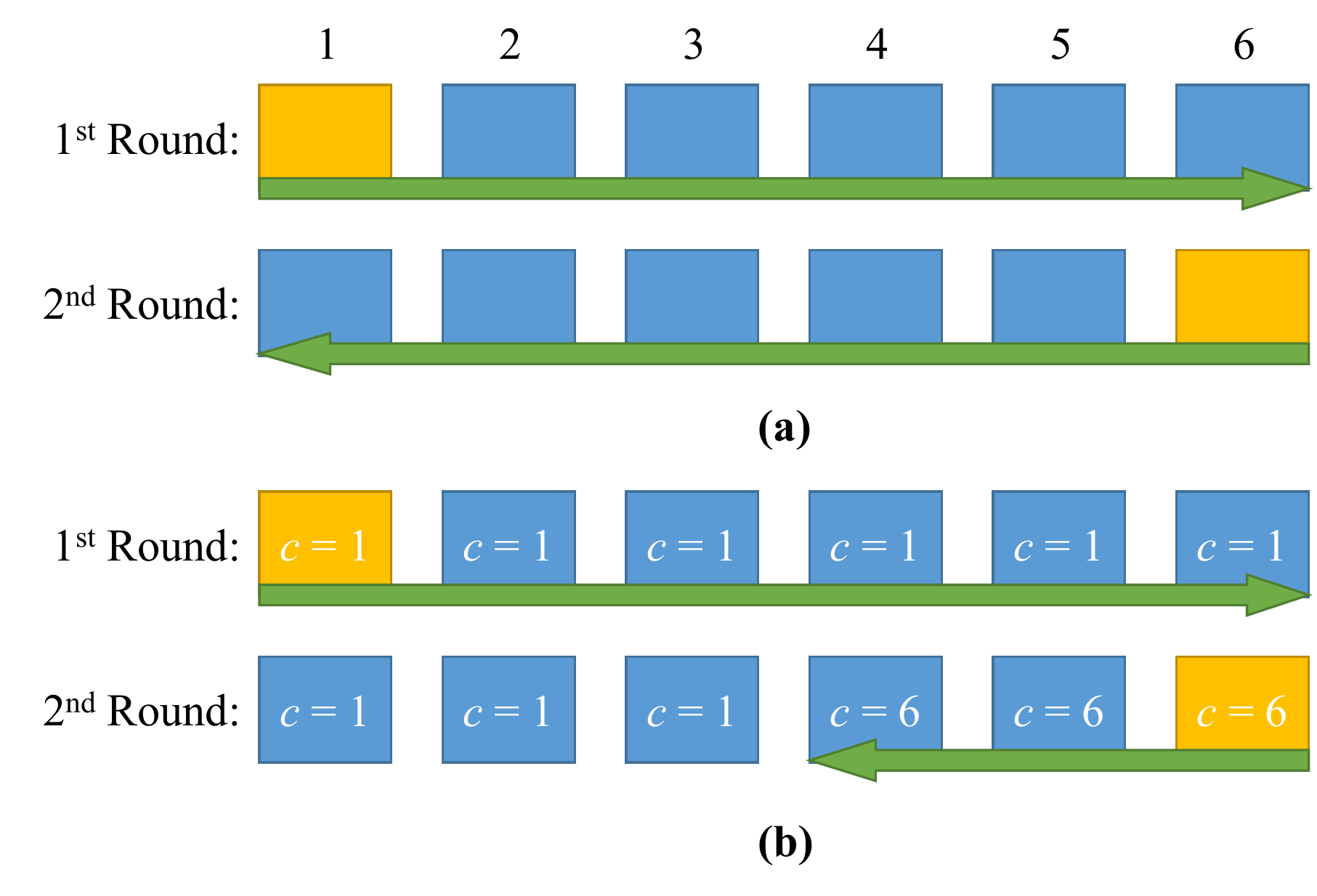}
	\caption{\small{Existing probability maps might be impaired during the iterative refinement. Therefore, we devise a checkpoint mechanism to avoid this issue. \textbf{Best viewed in color.}}}
	\label{fig:damage}
	\vspace{-12pt}
\end{figure}

\subsection{VS-ReID}
\label{sec:VS-ReID}
In this section, we will introduce the VS-ReID algorithm that combines previous two components to infer the masks of all instances on the whole video sequence.

As shown in Fig.~\ref{fig:pipeline}, given a video sequence and the mask (\ie ground-truth probability map) of the objects in the first frame, VS-ReID first initializes the probability maps $\{P\}$.
We enumerate all instances and forward propagate their probability maps from the first frame to the last frame by the mask propagation module.
After initialization, the re-identification module and mask propagation module are iteratively applied to the whole video sequence until no more high confidence instances can be found.
To be more specific, we first applied re-identification module to the whole video for all instances.
We keep the retrieved bounding box $\hat{x}$ with the highest similarity score $\hat{s}$.
Suppose $\hat{x}$ is the bounding box of instance $\hat{k}$ in frame $\hat{i}$, we then try to recover the probability map of instance $\hat{k}$ in frame $\hat{i}$, $P_{\hat{i},\hat{k}}$.
The recovery process is quite similar to the process of mask propagation, with one difference: there is no guided probability map from adjacent frames. 
So we replace that with the probability patch of instance $\hat{k}$ cropped from the first frame.
Once we obtain the recovered probability map, we can take it as the starting point and use the mask propagation module to bi-directionally recover more probability maps of instance $\hat{k}$ in adjacent frames.
However, sometimes existing probability maps will be impaired during this iterative refinement.
An example is shown in Fig.~\ref{fig:damage} (a), suppose we have 6 frames in a video sequence. 
In the first round of iterative refinement, we retrieve the instance $k$ in the first frame and propagate the recovered mask to the end of video sequence.
In the second round, we retrieve the instance $k$ again in the last frame and do the backward propagation.
In this case, all probability maps we predicted in the first round will be overwritten.
Because of the longer propagation distance, the probability map for instance $k$ in the second frame might be impaired in the second round.
To avoid this issue, we devise a checkpoint mechanism with a new variable $c_{i,k}$ recording the starting point by which $P_{i, k}$ is updated.
The initial value of $c_{i,k}$ is $1$, and every probability map prefers to be updated by a closer starting point.
As shown in Fig.~\ref{fig:damage} (b), the backward propagation will be interrupted at the fourth frame, since the first frame is closer to the third frame compared with the last one.
Finally, we combine all $\{P\}$ to generate the mask prediction $M$ through:
\[
	\small
	M_i(l) = \arg\underset{0 \le k \le K}{\max}\,\frac{1}{Z} *
	\begin{cases}
	P_{i,k}(l) &\mbox{$k \ne 0$}\\
	\prod_{j=1}^{K} (1-P_{i,j}(l)) &\mbox{$k = 0$}
	\end{cases}
\]
where $Z = \prod_{j=1}^{K} (1-P_{i,j}(l)) + \sum_{j=1}^{K} P_{i,j}(l)$ is a normalizing factor, $i$ is the frame index, $l$ is a pixel's location, $K$ is the number of instances in the video sequence. 

\subsection{Implementation Details}
\label{sec:implementation_details}
Two branches of mask propagation network are first trained individually.
The RGB branch is pre-trained on the MS-COCO~\cite{lin2014microsoft} and PASCAL VOC~\cite{everingham2010pascal} dataset.
During the pre-training, we use the randomly deformed ground-truth mask as the guided probability map.
Subsequently, the network is fine-tuned on the DAVIS training set.
The flow branch is initialized by RGB branch's weights and fine-tuned on the DAVIS training set.
Finally, those two branches are jointly fine-tuned together on the DAVIS training and validation sets.

Detection and re-identification networks are trained on the ImageNet~\cite{deng2009imagenet} dataset, we followed the training strategy in original papers~\cite{renNIPS15fasterrcnn, xiaoli2017joint}.
In particular, for the person category, we directly use the network in~\cite{xiaoli2017joint} as our re-identification network.

	%-------------------------------------------------------------------------

	%%%%%%%%% EXPERIMENTS
	\section{Experiments}
We evaluate our VS-ReID on the DAVIS 2017~\cite{Pont-Tuset_arXiv_2017} dataset.
DAVIS 2017 dataset contains 150 video sequences with all frames annotated with high-quality object masks.
There are 60 videos in the \emph{train set}, 30 videos in the \emph{val set}, 30 videos in the \emph{test-dev set} and 30 videos in the \emph{test-challenge set}.
In our experiments, we employ both \emph{train set} and \emph{val set} for training, and all performances are reported on the \emph{test-dev set}.
Followed~\cite{Perazzi2016}, we adopt region($\mathcal{J}$) and boundary($\mathcal{F}$) measures to evaluate the performance. 

\subsection{Ablation Study}
\begin{table}[t]
	\small
	\caption{Ablation study of each module in VS-ReID.}
	\centering
	\begin{tabular}{@{}l@{\,}|@{}c@{\,}|@{}c@{\,}|@{}c@{\,}|@{}c@{\,}}
		\hline
		&~$\mathcal{J}$-mean&~$\mathcal{F}$-mean~&~global-mean~&~boost~\\
		\hline\hline
		baseline\cite{Perazzi2017}~&0.509&0.526&0.517&-\\
		+~full-image to bbox~&0.532&0.577&0.555&~+~0.038\\
		+~flow-stream~&0.568&0.600&0.584&~+~0.007\\
		+~re-id module~&0.633&0.670&0.652&~+~0.068\\
		+~multi-scale testing~&0.644&0.678&\textbf{0.661}&~+~0.009\\
		\hline
	\end{tabular}
	\label{tab:ablation}
\end{table}
In this section, we investigate the effects of each component in VS-ReID model. 
Table~\ref{tab:ablation} summarizes how performance gets improved by adding each component step-by-step into our VS-ReID model.

We choose \cite{Perazzi2017} as our baseline model. 
After modified the input from full-image to bounding box, global-mean increases by $3.8\%$ and the boundary ($\mathcal{F}$) measure achieves a significant improvement of $5.1\%$.
It demonstrates that bounding box input overcomes large scale variations and contributes to capture the boundary details.
As mentioned in Sec.~\ref{sec:mask_propagation_module}, to incorporate the temporal information, we train an optical flow branch and joint fine-tuning it with the RGB branch.
This two-stream architecture also slightly improves the performance.
Employing the iterative refinement we introduced in Sec.~\ref{sec:VS-ReID} greatly improves the global-mean by $6.8\%$, which shows that the re-identification module and iterative refinement are essential.
We also visualize the example videos which are improved by this iterative refinement in Fig.~\ref{fig:visualization}.
Once an instance is recovered, it will benefit adjacent frames' prediction.
Finally, multi-scale testing further improves the results.

\begin{figure}
	\centering
	\includegraphics[width=0.48\textwidth]{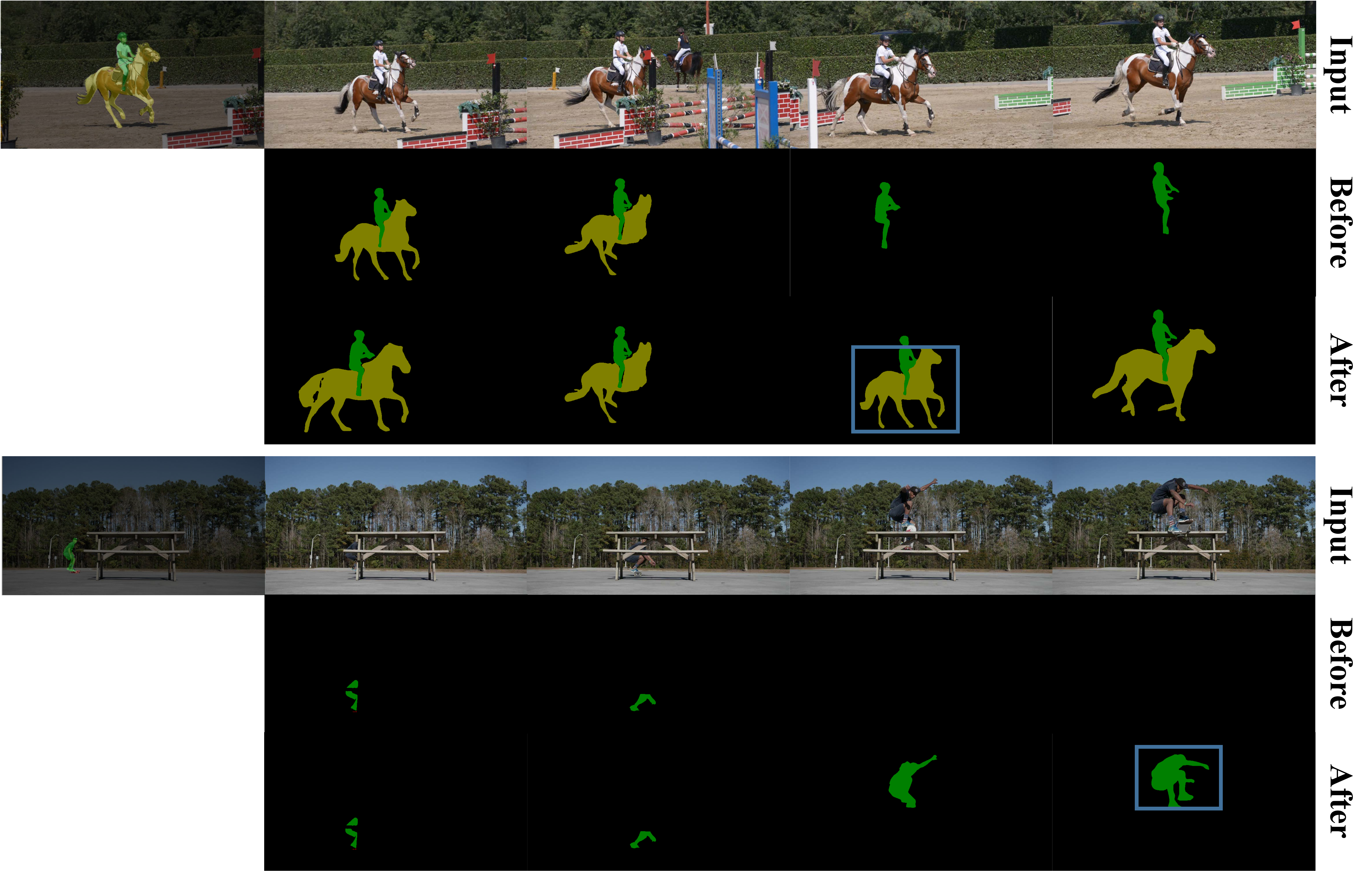}
	\caption{\small{Missing instances are retrieved by re-identification module. We annotate the retrieved instances by blue bounding boxes. \textbf{Best viewed in color.}}}
	\label{fig:reid}
	\vspace{-12pt}
\end{figure}

\begin{table*}[t]
	\caption{Results on 2017 DAVIS Challenge \textit{test-challenge set}.}
	\scriptsize
	\centering
	\begin{tabular}{@{}l@{\,}|@{\,}c@{ \,}@{\,}c@{ \,}@{\,}c@{ \,}@{\,}c@{ \,}@{\,}c@{ \,}@{\,}c@{ \,}@{\,}c@{ \,}@{\,}c@{ \,}@{\,}c@{ \,}@{\,}c@{ \,}@{\,}c@{ \,}@{\,}c@{ \,}@{\,}c@{ \,}@{\,}c@{ \,}@{\,}c@{ \,}@{\,}c@{ \,}@{\,}c@{ \,}@{\,}c@{ \,}@{\,}c@{ \,}@{\,}c@{ \,}@{\,}c@{ \,}@{\,}c@{ \,}}
		\hline
		Measure & Ours & Apata & Vanta & Haamo & Voigt & Lalal & Cjc & YXLKJ & Wasid & Froma & Zwrq0 & Drbea & Anews & Ilanv & Koh & Make & Kozab & Xn881 & Zpd & Griff & Nitin & Team5 \\
		\hline\hline
		Ranking & 1 & 2 & 3 & 4 & 5 & 6 & 7 & 8 & 9 & 10 & 11 & 12 & 13 & 14 & 15 & 16 & 17 & 18 & 19 & 20 & 21 & 22 \\
		\hline
		Global Mean$\uparrow$ & \textbf{69.9} & 67.8 & 63.8 & 61.5 & 57.7 & 56.9 & 56.9 & 55.8 & 54.8 & 53.9 & 53.6 & 51.9 & 50.9 & 49.7 & 49.1 & 48.0 & 47.8 & 47.6 & 47.1 & 42.0 & 25.6 & 11.2\\
		\hline\hline
		$\mathcal{J}$ Mean$\uparrow$ & \textbf{67.9} & 65.1 & 61.5 & 59.8 & 54.8 & 54.8 & 53.6 & 53.8 & 51.6 & 50.7 & 50.5 & 50.5 & 49.0 & 46.0 & 45.9 & 46.3 & 43.9 & 47.8 & 44.9 & 40.6 & 24.9 & 11.8\\
		$\mathcal{J}$ Recall$\uparrow$ & \textbf{74.6} & 72.5 & 68.6 & 71.0 & 60.8 & 60.7 & 59.5 & 60.1 & 56.3 & 54.9 & 54.9 & 56.4 & 55.1 & 49.3 & 50.2 & 50.0 & 45.8 & 56.3 & 48.0 & 42.1 & 12.3 & 7.3\\
		$\mathcal{J}$ Decay $\downarrow$ & 22.5 & 27.7 & 17.1 & 21.9 & 31.0 & 34.4 & 25.3 & 37.7 & 26.8 & 32.5 & 28.0 & 34.1 & 21.3 & 33.1 & 36.1 & 40.2 & 33.0 & 16.7 & 31.8 & 37.4 & 13.1 & 14.0\\
		\hline
		$\mathcal{F}$ Mean$\uparrow$ & \textbf{71.9} & 70.6 & 66.2 & 63.2 & 60.5 & 59.1 & 60.2 & 57.8 & 57.9 & 57.1 & 56.7 & 53.3 & 52.8 & 53.3 & 52.3 & 49.7 & 51.6 & 47.3 & 49.3 & 43.3 & 26.3 & 10.6\\
		$\mathcal{F}$ Recall$\uparrow$ & 79.1 & \textbf{79.8} & 79.0 & 74.6 & 67.2 & 66.7 & 67.9 & 62.1 & 64.8 & 63.2 & 63.5 & 57.9 & 58.3 & 58.4 & 57.1 & 52.8 & 56.0 & 53.0 & 54.4 & 43.2 & 9.1 & 3.0\\
		$\mathcal{F}$ Decay$\downarrow$ & 24.1 & 30.2 & 17.6 & 23.7 & 34.7 & 36.1 & 27.6 & 42.9 & 28.8 & 33.7 & 30.4 & 39.5 & 23.7 & 36.4 & 39.2 & 44.8 & 36.3 & 21.6 & 36.2 & 40.2 & 13.0 & 12.6\\
		\hline
	\end{tabular}
	\label{tab:leader_board}
\end{table*}

\subsection{Benchmark}
\begin{figure*}[t]
	\centering
	\includegraphics[width=1.0\textwidth]{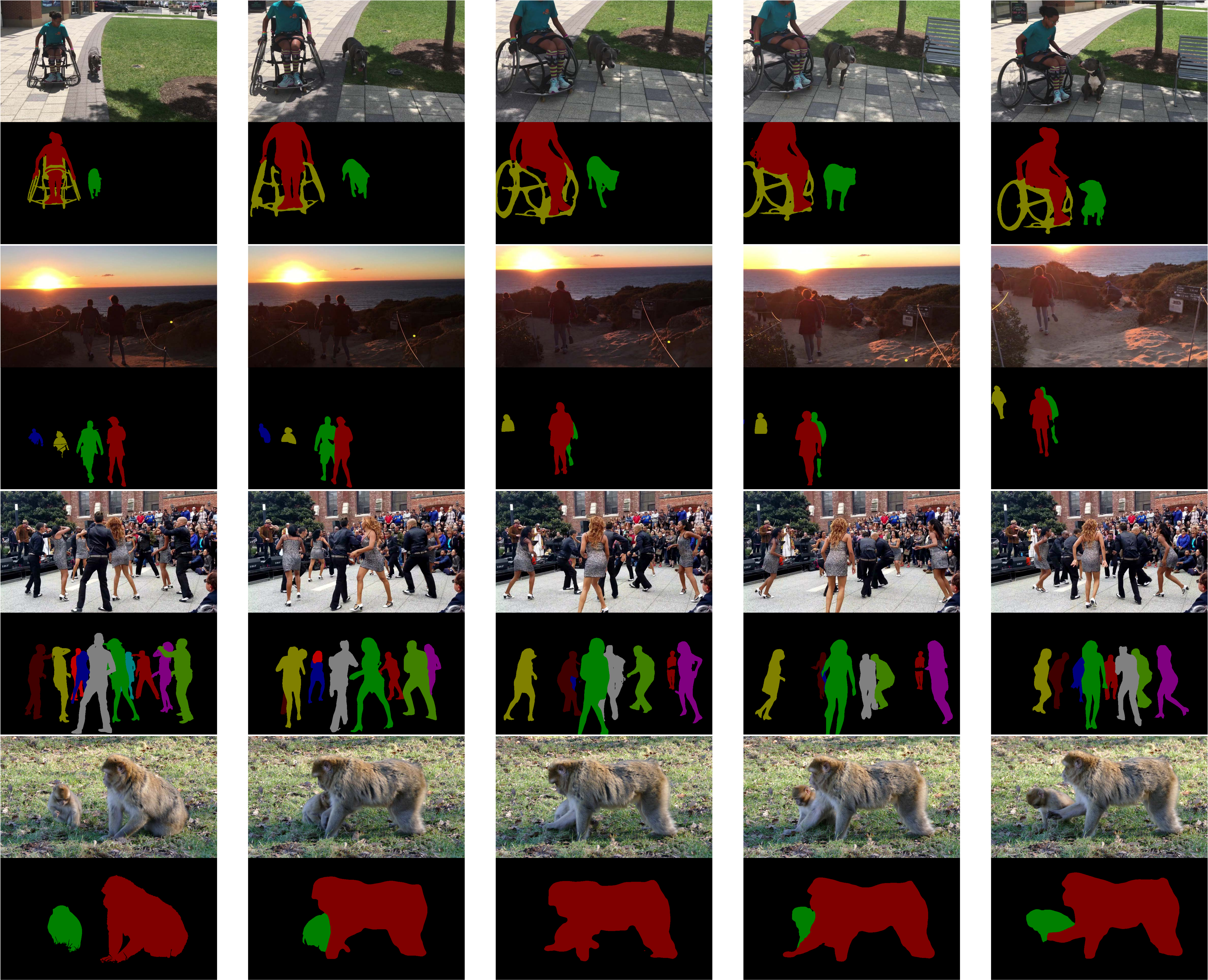}
	\caption{\small{Qualitative results of our VS-ReID model on DAVIS 2017 \textit{test-dev set} and \textit{test-challenge set}.}}
	\label{fig:visualization}
	\vspace{-12pt}
\end{figure*}

As shown in Table~\ref{tab:leader_board}, VS-ReID achieves a global mean of 0.699 on \emph{test-challenge set}, the best performance in 2017 DAVIS Challenge. 
By inspecting closer, we observe that VS-ReID wins both $\mathcal{J}$-Mean and $\mathcal{F}$-Mean measures and outperforms the second place method by more than $2\%$.
Thanks to the re-identification module that incorporates the long-term memory, our $\mathcal{J}$-Decay and $\mathcal{F}$-Decay are also relatively small.
In Fig.~\ref{fig:visualization}, we demonstrate some examples of VS-ReID prediction on DAVIS \emph{test-dev set} and \emph{test-challenge set}.

	%-------------------------------------------------------------------------

	%%%%%%%%% CONCLUSION
	
\section{Conclusion}

In this work we tackle the problem of video object segmentation and explore the utility of object re-identification. 
We propose Video Object Segmentation with Re-identification (VS-ReID) model which includes two dedicated modules: a mask propagation module and a ReID module.
It is observed that our ReID module combined with bidirectional refinement is capable of retrieving missing instances and greatly improves the performance.
These two modules are employed iteratively, enabling our final model to win the DAVIS video segmentation challenge.

	%-------------------------------------------------------------------------
	
	{\small
		\bibliographystyle{ieee}
		\bibliography{egbib}
	}
	
\end{document}